\documentclass[letterpaper, 10 pt, conference]{ieeeconf}  

\IEEEoverridecommandlockouts                              

\title{\LARGE \bf
Enhancing Context-Aware Human Motion Prediction for \\Efficient Robot Handovers}

\author{Gerard Gómez-Izquierdo$^{1}$, Javier Laplaza$^{1}$, Alberto Sanfeliu$^{1}$ and Anaís Garrell$^{1}$} 

\usepackage{graphicx} 
\usepackage{hyperref} 
\usepackage{amssymb}
\usepackage{array}
\usepackage{tabularx}
\usepackage{amsmath}
\usepackage{dblfloatfix}
\usepackage{comment}

\usepackage{balance}

\begin{document}

\maketitle
\thispagestyle{empty}
\pagestyle{empty}

\begin{abstract}


Accurate human motion prediction (HMP) is critical for seamless human-robot collaboration, particularly in handover tasks that require real-time adaptability. Despite the high accuracy of state-of-the-art models, their computational complexity limits practical deployment in real-world robotic applications. In this work, we enhance human motion forecasting for handover tasks by leveraging \textit{siMLPe} \cite{guo2023back}, a lightweight yet powerful architecture, and introducing key improvements. Our approach, named \textit{IntentMotion} incorporates intention-aware conditioning, task-specific loss functions, and a novel intention classifier, significantly improving motion prediction accuracy while maintaining efficiency. Experimental results demonstrate that our method reduces body loss error by over 50\%, achieves 200× faster inference, and requires only 3\% of the parameters compared to existing state-of-the-art HMP models. These advancements establish our framework as a highly efficient and scalable solution for real-time human-robot interaction.

\end{abstract}


\section{INTRODUCTION}

Human motion prediction (HMP) plays a crucial role in human-robot collaboration (HRC) by enabling robots to anticipate human movements and respond proactively. This capability is particularly important in handover tasks, where the seamless exchange of objects between humans and robots requires both accuracy and speed. The ability to predict human motion allows robots to preemptively adjust their trajectories, improving efficiency and ensuring safety. In this context, human intention—whether the motion is collaborative or non-collaborative—directly influences the prediction and subsequent robot response. As a result, HMP is a critical component in various real-world robotic applications, including industrial automation, assistive robotics, and service robots.

Significant progress has been made in HMP through state-of-the-art (SOTA) deep learning approaches, which leverage architectures such as recurrent neural networks (RNNs), graph convolutional networks (GCNs), and transformers. These models effectively capture both the temporal and spatial dependencies of human motion, leading to highly accurate predictions. However, their computational complexity remains a key limitation. Many high-accuracy models rely on millions of parameters and require substantial inference times, which hinders their deployment in real-time robotic applications where rapid decision-making is essential.

An important yet less explored challenge in HMP is the trade-off between model complexity and prediction accuracy. Many existing methods prioritize precision but overlook computational feasibility, making them impractical for real-world robotic systems. Furthermore, while some approaches integrate contextual information, such as environmental constraints, few explicitly incorporate human intention into the prediction pipeline. Given that human motion is inherently influenced by intent, recognizing these underlying factors, collaborative or non-collaborative, before forecasting full-body movement could enhance accuracy and lead to more natural and effective human-robot interactions.

\begin{figure}[t]
    \centering
\includegraphics[width=1\linewidth]{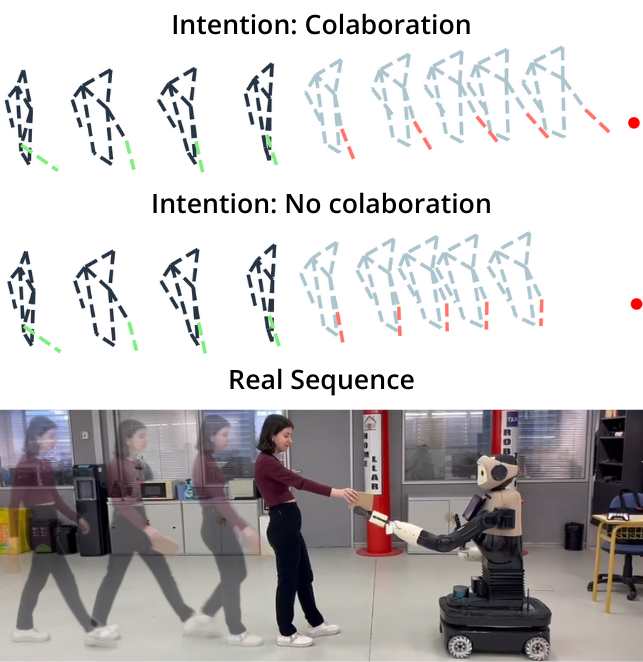}
    \caption{\textbf{Motion prediction based on detected intention.} Both sequences share the same ground truth represented by a dark blue skeleton and a green right hand. \textbf{Top:} Prediction with collaborative intention. \textbf{Middle:} Prediction with non-collaborative intention. \textbf{Red Circle:} Represents the last frame of the robot's end effector. \textbf{Bottom:} Real handover sequence.}\vspace{-\baselineskip}
    \label{fig:figure1}
\end{figure}


This work addresses these challenges by leveraging \textit{siMLPe}, a recently proposed lightweight architecture for human motion prediction, and adapting it specifically for human-robot handover tasks. Our approach, named \textit{IntentMotion}, introduces key enhancements, including task-specific loss functions tailored for handover scenarios, an intention-aware conditioning mechanism, and a novel intention classifier that refines motion predictions. The evaluation of our system is performed in a handover task setting, where the robot predicts the motion of the human while classifying the intention as collaborative or non-collaborative. Through extensive experimentation, we demonstrate that our method achieves high prediction accuracy while significantly reducing computational requirements, making it a practical and efficient solution for real-time human-robot collaboration.

The remainder of the paper is organized as follows. In Section II, we present the related work. The methodology is specified in Section III. In Section IV, we describe the experimentation, which is subsequently analyzed in the results and discussion in Section V. Finally, conclusions are given in Section VI.

\section{Related Work}

Human motion prediction plays a crucial role in human-robot interaction (HRI), enabling robots to anticipate human movements and respond proactively. Several studies have explored this problem in collaborative tasks, focusing on improving the accuracy and efficiency of motion forecasting. Liu and Wang \cite{liu2017human} introduced a predictive framework for collaborative manufacturing, allowing robots to anticipate human motion. Zhao et al. \cite{zhao2020experimental} performed an evaluation of human motion prediction models in collaborative settings, highlighting the advantages of parameter-efficient neural network architectures. Mugisha et al. \cite{mugisha2024motion} applied Gaussian Process models to predict human hand motion, leveraging gaze and hand movement information to enhance intention detection.

A particularly relevant work is that of Laplaza et al. \cite{laplaza2022iros, laplaza2022context}, which focuses on predicting human motion in handover tasks. Their approach builds upon the architecture proposed by Mao et al. \cite{mao2020history} and incorporates human intention as a conditioning factor for motion predictions. The model was trained and validated on a custom dataset of human-to-robot handovers. In our work, we employ the same dataset to assess improvements in computational efficiency and prediction accuracy over Laplaza et al.’s method.

Over the years, various deep learning architectures have been employed to improve human motion prediction. Recurrent Neural Networks (RNNs) have been widely used due to their ability to model sequential data.In \cite{fragkiadaki2015recurrent, carreira2016human} the authors proposed an encoder-decoder model based on Long Short-Term Memory (LSTM) networks to capture motion dynamics. Martinez et al. \cite{martinez2017human} improved this approach by introducing residual connections that helped model motion velocities, leading to smoother predictions. However, RNN-based models suffer from issues such as vanishing gradients and difficulties in capturing long-term dependencies, which limit their effectiveness in complex motion forecasting tasks.

To better capture spatial relationships between human joints, Graph Convolutional Networks (GCNs) have been explored as an alternative. Mao et al. \cite{mao2020history} introduced a GCN-based model that encodes both spatial and temporal dependencies, significantly improving prediction accuracy. Mao et al. \cite{mao2021multi} further refined this approach by incorporating an attention mechanism that selectively emphasizes relevant motion patterns from different body parts based on past observations. Other works, such as Ma et al. \cite{ma2022progressively}, extended GCNs by introducing multiscale representations to enhance long-term prediction accuracy.

With the growing popularity of attention mechanisms, Transformer-based models have also been explored for human motion prediction. Aksan et al. \cite{aksan2021spatio} developed a spatio-temporal Transformer that jointly models spatial and temporal dependencies to enhance motion forecasting. Mao et al. \cite{mao2021multi} introduced a memory-augmented Transformer that retrieves historical motion patterns to guide predictions. While these methods achieve state-of-the-art accuracy, they require a large number of parameters, making them computationally expensive and challenging to deploy in real-time applications.

A recent breakthrough in human motion prediction is the introduction of \textit{siMLPe}, a lightweight yet powerful MLP-based model proposed by Guo et al. \cite{guo2023back}. Unlike deep GCNs or Transformer-based architectures, \textit{siMLPe} applies a Discrete Cosine Transform (DCT) to encode temporal dependencies and predicts motion using fully connected layers with layer normalization. Despite its simplicity, this model outperforms previous architectures on major benchmarks while maintaining an extremely low computational footprint. Its efficiency and accuracy make it particularly well-suited for real-time robotic applications.

The present work builds upon previous research by adapting \textit{siMLPe} for human-robot handover tasks and integrating intention-aware conditioning to enhance motion prediction accuracy and computational efficiency.

\section{METHODOLOGY}

In this section, we present our approach for intention-aware human motion prediction in handover tasks. We enhance \textit{siMLPe} \cite{guo2023back} by introducing \textit{IntentMotion}, our intention-aware motion prediction framework.

Here,  first define the problem formulation, detailing how human motion sequences and intention labels are represented. Next, we describe the architecture of our proposed \textit{IntentMotion} framework, which extends prior frequency-domain motion prediction models by incorporating intention as contextual information. We then introduce key modifications, including an intention embedding mechanism and novel loss functions designed to enhance the accuracy of right-hand motion. Finally, we outline the design and training objectives of our intention classifier, which enables real-time intent recognition to dynamically refine motion forecasts.

\begin{figure*}[t]
    \centering
    \includegraphics[width=0.95\linewidth]{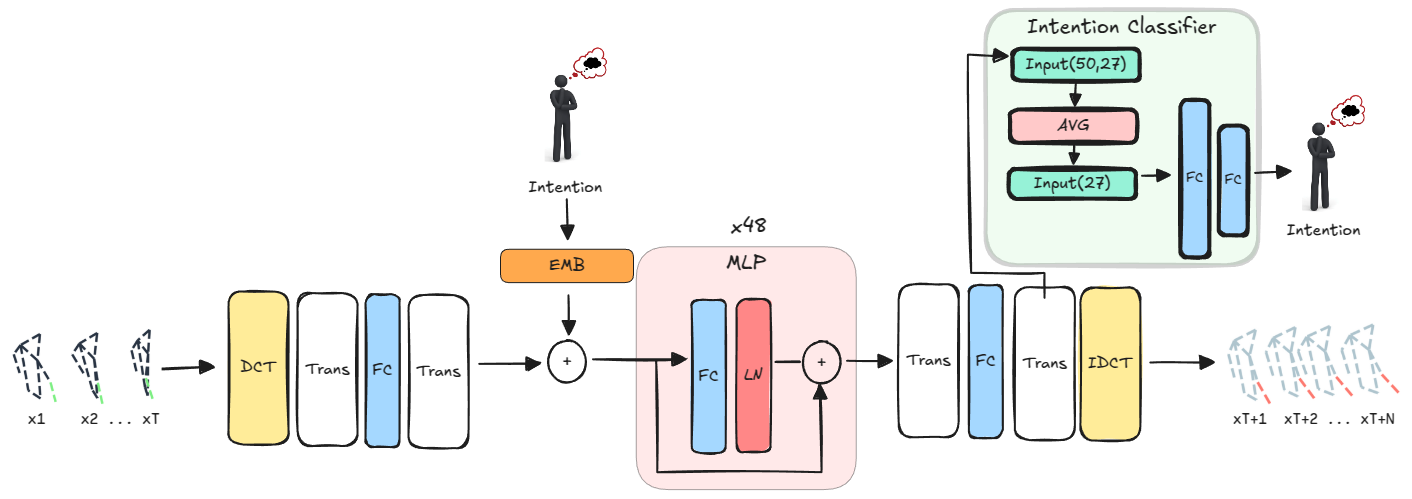}
    \caption{\textbf{Overview of our \textit{siMLPe}-based approach for human motion prediction.} \textit{FC} denotes a fully connected layer, \textit{LN} represents layer normalization, \textit{Trans} indicates a transpose operation, \textit{EMB} denotes the embedding layer, and \textit{AVG} represents the average pooling layer. \textit{DCT} and \textit{IDCT} correspond to the discrete cosine and inverse discrete cosine transformations, respectively. The MLP module (highlighted in pink) consists of 48 \textit{FC} and \textit{LN} layers repeated across the architecture. The green box represents how the intention classifier processes data before intention prediction.}\vspace{-\baselineskip}
    \label{fig:arch_comb} 
\end{figure*}

\subsection{Problem Definition}

Given a sequence of past 3D human poses, the objective is to accurately forecast the corresponding future motion. The observed human motion sequence is represented as  
\(\mathbf{x}_{1:T} = \left[ \mathbf{x}_1, \dots, \mathbf{x}_T \right]\in \mathbb{R}^{T \times C}\),  
where each pose \(\mathbf{x}_t \in \mathbb{R}^{C}\) at time step \(t\) is encoded as a \(C\)-dimensional vector. Specifically, \(\mathbf{x}_t\) contains the 3D coordinates of body joints, with \(C = 3 \times K\), where \(K\) denotes the total number of joints. The goal is to predict the sequence of future motion frames, formulated as  
\(\mathbf{x}_{T+1:T+N} = \left[ \mathbf{x}_{T+1}, \dots, \mathbf{x}_{T+N} \right] \in \mathbb{R}^{N \times C}\).  

Additionally, each input sequence \(\mathbf{X}_{1:T}\) is associated with an intention label \( i \in \{0, 1\} \), where \(i = 0\) corresponds to a collaborative intention and \(i = 1\) to a non-collaborative intention. This intention variable encodes the expected human intent in the predicted motion frames, denoted as \(\mathbf{\hat{i}}_{N+1:N+T}\), providing contextual information to enhance motion forecasting accuracy.

\subsection{Network Overview} 

\textit{IntentMotion} architecture, as done in \cite{guo2023back} operates in the frequency domain by applying the Discrete Cosine Transform (DCT) to encode temporal dependencies, followed by fully connected layers, transpose operations, and layer normalization. The output is then transformed back to the original pose representation using the Inverse Discrete Cosine Transform (IDCT). Given an input sequence of past 3D human poses
\(\ \mathbf{x}_{1:T} = \left[ \mathbf{x}_1, \dots, \mathbf{x}_T \right] \in \mathbb{R}^{T \times C} \)
the model predicts the sequence of future poses
\(\ \mathbf{x}'_{T+1:T+N} = \left[ \mathbf{x}'_{T+1}, \dots, \mathbf{x}'_{T+N} \right] \in \mathbb{R}^{N \times C}. \) The prediction process can be formulated as:
\(\ \hat{\mathbf{x}}_{T+1:T+N} = \mathcal{D}^{-1} \big( \mathcal{F} (\mathcal{D} (\mathbf{x}_{1:T})) \big), \)
where \(\mathcal{F}\) denotes the network function, \(\mathcal{D}\) is the DCT transformation, and \(\mathcal{D}^{-1}\) represents the IDCT transformation.

\subsection{Proposed Modifications}

Figure \ref{fig:arch_comb} shows the architecture of our network. 
\subsubsection{Incorporating Intention as Context}
To integrate intention into the model, we encode the intention label \( i \in \{0,1\} \) using an embedding layer:
\begin{equation}
    \mathbf{e}_i = \text{Embedding}(i), \quad \mathbf{e}_i \in \mathbb{R}^{1 \times C}.
\end{equation}

The embedded intention vector is then expanded along the temporal dimension to match the sequence length \(T\), forming the intention matrix:
\begin{equation}
    \mathbf{M}_i = \mathbf{e}_i \otimes \mathbf{1}_{T}, \quad \mathbf{M}_i \in \mathbb{R}^{T \times C},
\end{equation} 
where \(\otimes\) denotes broadcasting across the temporal dimension.

Next, we fuse the encoded intention with the transformed motion sequence \( \mathbf{z}^0 \), obtained after the first fully connected layer:
\begin{equation}
    \mathbf{z}^0 = \mathcal{D}(\mathbf{x}_{1:T}) \mathbf{W}_0 + \mathbf{b}_0, 
\end{equation} 
where \(\mathbf{W}_0 \in \mathbb{R}^{C \times C}\) and \(\mathbf{b}_0 \in \mathbb{R}^{C}\) are learnable parameters. The fusion is achieved through element-wise addition:
\begin{equation}
    \mathbf{U} = \mathbf{M}_i + \mathbf{z}^0. 
\end{equation} 
This enriched representation \( \mathbf{U} \) encodes both the historical motion and the expected human intention, enabling the model to generate predictions that are contextually aware of the intention label.

\subsubsection{Objective Function}

The objective function of our model is designed to improve the predictive performance by incorporating additional loss terms that explicitly account for the movement dynamics of the right hand during handover tasks. Our total loss function, \( \mathcal{L}_h \), extends the loss formulation of \textit{siMLPe} \cite{guo2023back}, which originally includes the terms \( \mathcal{L}_{re} \) and \( \mathcal{L}_{v} \), by adding three additional terms: \( \mathcal{L}_c \), \( \mathcal{L}_{rer} \), and \( \mathcal{L}_{vr} \). The final objective function is given by:
\begin{equation}
    \mathcal{L}_h = \mathcal{L}_{re} + \mathcal{L}_{v} + \mathcal{L}_{c} + \mathcal{L}_{rer} + \mathcal{L}_{vr}.
\end{equation}

The standard reconstruction loss \( \mathcal{L}_{re} \) and velocity loss \( \mathcal{L}_{v} \) are defined as:
\begin{equation}
    \mathcal{L}_{re} = \| \mathbf{x}'_{T+1:T+N} - \mathbf{x}_{T+1:T+N} \|_2^2,
\end{equation}
\vspace{-13pt}
\begin{equation}
    \mathcal{L}_{v} = \| \mathbf{v}'_{T+1:T+N} - \mathbf{v}_{T+1:T+N} \|_2^2,
\end{equation}
    
where \( \mathbf{x}' \) and \( \mathbf{x} \) denote the predicted and ground-truth motion sequences, respectively, and \( \mathbf{v} \) represents the velocity, computed as \( \mathbf{v}_t = \mathbf{x}_{t+1} - \mathbf{x}_{t} \).

To overcome the limitation of \textit{siMLPe}, which tends to predict static right-hand poses, we introduce an additional loss term \( \mathcal{L}_{c} \) that ensures the right hand moves towards the robot end effector in collaborative scenarios. This term consists of two sub-components: \( \mathcal{L}_{r} \) and \( \mathcal{L}_{b} \):
\begin{equation}
    \mathcal{L}_{c} = 0.05 \mathcal{L}_{r} + 0.95 \mathcal{L}_{b}.
\end{equation}
    
Here, \( \mathcal{L}_{r} \) minimizes the Euclidean distance between the last predicted right hand position \( \mathbf{x}'_{rh, T+N} \) and the robot end effector \( \mathbf{ree}_{T+N} \):
\begin{equation}
    \mathcal{L}_{r} = \| \mathbf{x}'_{rh, T+N} - \mathbf{ree}_{T+N} \|_2^2.
\end{equation}

Additionally, \( \mathcal{L}_{b} \) penalizes deviations from the natural movement constraints of the right hand by minimizing the difference in average distance between the right hand and \( K \) other body joints in the predicted and ground-truth sequences:
\begin{equation}
    \mathcal{L}_{b} = \left\| \frac{1}{K} \sum_{k=1}^{K} \| \mathbf{x}'_{rh} - \mathbf{x}'_k \|_2 - \frac{1}{K} \sum_{k=1}^{K} \| \mathbf{x}_{rh} - \mathbf{x}_k \|_2 \right\|_2^2.
\end{equation}

This term enforces consistency in the natural movement patterns of the right hand relative to the rest of the body.

To further guide the prediction of right-hand motion, we introduce \( \mathcal{L}_{rer} \) and \( \mathcal{L}_{vr} \), which emphasize minimizing positional and velocity errors of the right hand. The error for the right hand, \( \mathcal{L}_{rer} \), is formulated as:
\begin{equation}
    \mathcal{L}_{rer} = \| \mathbf{x}'_{rh, T+1:T+N} - \mathbf{x}_{rh, T+1:T+N} \|_2^2.    
\end{equation}
    
Similarly, the velocity consistency loss \( \mathcal{L}_{vr} \) is given by:
\begin{equation}
    \mathcal{L}_{vr} = \| \mathbf{v}'_{rh, T+1:T+N} - \mathbf{v}_{rh, T+1:T+N} \|_2^2,
\end{equation}
    
where \( \mathbf{v}'_{rh} \) and \( \mathbf{v}_{rh} \) represent the predicted and ground-truth velocities of the right hand, respectively.

It is important to note that the loss term \( \mathcal{L}_{c} \) is applied selectively, being computed only for sequences labeled with collaborative intent. This ensures that the model learns context-specific behavior without introducing biases in non-collaborative scenarios.

\subsection{Proposed Intention Classifier}

To enable automatic conditioning of the motion predictor with the human's intention in real-world scenarios, we propose an intention classifier that infers intention labels directly from body keypoint trajectories. This classifier is designed to be lightweight and efficient, drawing inspiration from \cite{guo2023back} while adapting its structure for classification tasks. Figure \ref{fig:arch_comb} also shows the architecture of the classifier.

\subsubsection{Architecture}
The classifier adopts a similar backbone to \textit{IntentMotion} but omits the intention embedding module. Instead, the output feature map from the last fully connected layer is aggregated across the temporal dimension via average pooling. Let \(\mathbf{F} \in \mathbb{R}^{T \times C}\) denote the feature representation extracted from an input sequence of \(T\) frames, where \(C\) is the feature dimension. The pooled feature is computed as:
\begin{equation}
    \bar{\mathbf{F}} = \frac{1}{T} \sum_{t=1}^{T} \mathbf{F}_t.
\end{equation}

This pooled feature is subsequently mapped to class logits through a series of fully connected layers with nonlinear activations. Formally, the classifier computes:
\begin{equation}
    \mathbf{z} = g(\bar{\mathbf{F}}) \in \mathbb{R}^{M},
\end{equation}

where \(g(\cdot)\) represents the mapping function composed of two fully connected layers, and \(M\) is the number of intention classes (in our experiments, \(M=2\)). The predicted intention is then given by:
\begin{equation}
    \hat{y} = \operatorname{argmax}(\mathbf{z}).
\end{equation}

\subsubsection{Objective Function}
The training objective for the classifier modifies the loss formulation of the motion predictor. In addition to the \(\mathcal{L}_{re}\) loss and the velocity loss \(\mathcal{L}_{v}\) already defined in \cite{guo2023back}, we incorporate a cross-entropy loss \(\mathcal{L}_{ce}\) to supervise the intention predictions. The overall objective is formulated as:
\begin{equation}
    \mathcal{L}_{class} = \mathcal{L}_{re} + \mathcal{L}_{v} + \mathcal{L}_{ce}.
\end{equation}

Here, the cross-entropy loss is defined by:
\begin{equation}
    \mathcal{L}_{ce} = -\sum_{i=1}^{M} y_i \log \left(\frac{\exp(z_i)}{\sum_{j=1}^{M}\exp(z_j)}\right),
\end{equation}

where \(y_i\) is the ground-truth label for class \(i\) and \(z_i\) denotes the corresponding logit. This formulation ensures that the classifier performs classification capturing the spatio-temporal characteristics of the motion to provide reliable intention cues to condition the motion predictions.

\begin{table}[b]
\vspace{-10pt}
\begin{center}
\begin{tabular}{| >{\centering\arraybackslash}m{8em} | >{\centering\arraybackslash}m{0.5cm} | >{\centering\arraybackslash}m{1cm} | >{\centering\arraybackslash}m{1cm} | >{\centering\arraybackslash}m{0.5cm} |}
\hline
\textbf{Model} & Body $L_2$ (m) & $\leq$0.35m (\%) & $\leq$0.40m (\%) & Right Hand $L_2$ (m) \\
\hline
RNN \cite{martinez2017human} & 0.739 & 3.49 & 11.62 & 0.677 \\
\hline
Hist. Rep. Its.  \cite{mao2020history} & 0.403 & 34.13 & 37.14 & 0.188 \\
\hline
Laplaza et al. \cite{laplaza2022iros} & 0.355& 32.15 & 35.71 & \textbf{0.151} \\
\hline
siMLPe \cite{guo2023back} & \textbf{0.177}& \textbf{96.95} & \textbf{100.00} & 0.217 \\
\hline
\end{tabular}
\caption{Comparison of Original siMLPe with previous models}\vspace{-\baselineskip}
\label{table_comparison_others}
\end{center}
\end{table}
\section{Experimentation}

\subsection{Dataset}
We adopt a dataset originally detailed by Laplaza et al. \cite{laplaza2022iros} to train and evaluate our approach. In this dataset, human–robot handover interactions are captured in controlled experiments using the anthropomorphic robot IVO. During each trial, a human volunteer, acting as the giver, transfers a 10 cm cylindrical object to the robot, which functions as the receiver by aligning its motion to successfully grasp the object. All interactions commence from a 6-meter separation and conclude upon completion of the transfer.

\begin{figure*}[t]
    \centering
    \includegraphics[width=1\linewidth]{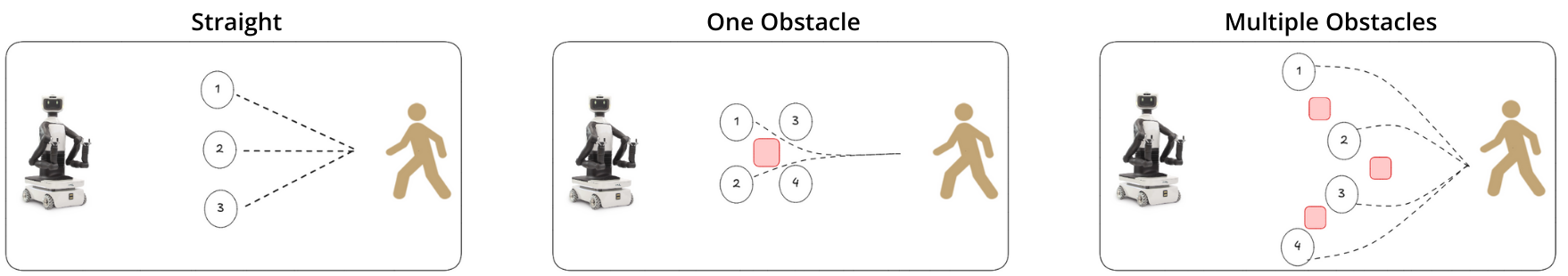}
    \caption{\textbf{The figure presents three top-view scenarios from the dataset.} The human (right figure) and the robot (left figure) interact in different environments, with obstacles depicted as red squares. Dashed lines indicate human movement, while the robot follows corresponding target points.} \vspace{-\baselineskip}
    \label{fig:laplaza_dataset}
\end{figure*}

The dataset encompasses multiple scenarios: one without obstacles, another with a single obstacle, and a final setup featuring three obstacles. Additionally, various approach paths were defined to capture both lateral and straight trajectories, as shown in Figure \ref{fig:laplaza_dataset}. Volunteers executed each trajectory under three behavioral conditions: a natural execution, a gesture-modulated variant, and an adversarial variant. For our purposes, the original four intention labels are consolidated into two classes. Specifically, both natural and gesture-modulated behaviors are treated as \emph{collaboration}, while neutral and adversarial behaviors are grouped as \emph{no collaboration}.

\subsection{Training Details}

Both \textit{IntentMotion} and \textit{IntentMotion Classifier} are trained using sequences of 50 frames as input, with the subsequent 10 frames (corresponding to 1 second of data) serving as the target output. To extend the prediction horizon to 2.5 seconds, the models are deployed in an autoregressive way. In the case of the intention classifier, the final predicted intention is determined by taking the mode across three consecutive blocks of 10 predicted frames.

Models are trained for 5000 epochs using the Adam optimizer combined with a cosine annealing scheduler, which adjusts the learning rate from \(10^{-2}\) down to \(10^{-5}\). A batch size of 256 is used during training. The feature dimension is defined as \(C = 3 \times K\), where \(K = 9\) represents the number of upper-body keypoints extracted from the handover data.

In addition, data augmentation is employed by randomly inverting the motion sequence during training, similar to the procedure in \cite{guo2023back}. An extensive ablation study was conducted to fine-tune all model hyperparameters, thereby ensuring optimal performance across both tasks.

\begin{table}[b]
\vspace{-15pt}
\begin{center}
\begin{tabular}{|c|c|c|}
\hline
\textbf{Model} & \textbf{Mean inference time (ms)} & \textbf{\# Parameters} \\
\hline
Laplaza \cite{laplaza2022iros} + int. & 3,680.6 & 6,113,782 \\
\hline
IntentMotion & 18.5 & 126,558 \\
\hline
IntentMotion Classifier & 18.7 & 265,032 \\
\hline
\end{tabular}
\caption{Comparison of Inference Time and Parameters}\vspace{-\baselineskip}
\label{table_efficiency_comparison}
\end{center}
\end{table}

\subsection{Evaluation Details}

To evaluate \textit{IntentMotion}, we employ metrics analogous to those by Laplaza et al. \cite{laplaza2022iros}. The primary evaluation metric is the mean L2 distance between the predicted and ground truth upper-body 3D joint positions, computed for each frame. 

To facilitate a more detailed analysis, we report the percentage of predicted frames with errors below predefined thresholds of 0.35m and 0.40m, metrics that directly enable comparison with previous approaches (see Table \ref{table_comparison_others}). In our ablation study shown in Table \ref{table_ablation}, we further introduce thresholds of 0.30m and 0.20m, given that our best model nearly achieves 100\% accuracy under the 0.35m and 0.40m criteria. 

Moreover, since the original \textit{siMLPe} \cite{guo2023back} architecture tends to produce static predictions for the right-hand, which is critical in handover tasks, we incorporate a dedicated L2 error metric for the right-hand joint to quantify improvements in its motion prediction (see Table \ref{table_ablation}).

For the \textit{IntentMotion Classifier}, our evaluation is centered on the macro F1-score. This metric is well-suited to handle the inherent class imbalance in our dataset, where collaborative samples predominate over non-collaborative ones.

To robustly assess both models, as done in \cite{laplaza2022iros} we adopt a leave-one-out cross-validation strategy over 10 subjects, where each subject is sequentially held out for testing while the remaining subjects form the training set. This approach ensures a comprehensive evaluation of model generalization across different individuals.

Finally, we evaluate the computational efficiency of our models by measuring the average inference time (in milliseconds over 100 inferences) and by comparing the number of model parameters with those  in \cite{laplaza2022iros} (see Table \ref{table_efficiency_comparison}).

\section{Results and Discussion}
This section presents a comprehensive evaluation of our proposed \textit{IntentMotion} framework, analyzing both quantitative performance metrics and qualitative insights. We first compare our model against existing baselines in terms of computational efficiency and motion prediction accuracy, emphasizing the improvements introduced by intention conditioning. We then conduct ablation studies to assess the contributions of individual loss functions and intention-aware mechanisms. Additionally, we evaluate the performance of the \textit{IntentMotion Classifier} in recognizing human intent, highlighting its impact on motion prediction. Finally, qualitative results provide visual evidence of how our approach refines human-robot handover interactions, demonstrating its ability to adapt motion predictions based on inferred intention.

\balance
\subsection{Quantitative Metrics}

As shown in Table \ref{table_efficiency_comparison}, our proposed model achieves a substantial reduction in both inference time and parameter count compared to the baseline in \cite{laplaza2022iros}, requiring only ~3\% of the parameters and achieving 200× faster inference. This gain in computational efficiency is particularly valuable in robotic scenarios, where multiple specialized models may need to operate in real time to support robust human–robot collaboration.

Moreover, as shown in Table \ref{table_comparison_others} our approach significantly lowers the L\(_2\) error for the entire body, improving from 0.355m to 0.165m relative to \cite{laplaza2022iros}. Such accuracy is essential for safe and precise interactions in collaborative tasks, where even minor positional inaccuracies can undermine human comfort or safety.

The ablation results in Table \ref{table_ablation} further highlight the benefits of incorporating intention conditioning. When the ground-truth intention label is provided, every performance metric exceeds that of the finetuned \textit{IntentMotion} baseline, demonstrating how context awareness enhances predictive fidelity and reinforcing the importance of a reliable intention classifier.

Because handover tasks require highly accurate right-hand positioning, we additionally measure the L\(_2\) distance for the right-hand joint. This evaluation underscores the limitations of the original \textit{siMLPe} framework, which often produces static right-hand poses. By contrast, our model, through new loss terms and refined training, avoids static right hand predictions and reduces the joint-level error from 0.217m to 0.195m (see Table \ref{table_ablation}).

Table \ref{table_ablation} also illustrates how the introduction of \(\mathcal{L}_{c}\), \(\mathcal{L}_{rer}\), and \(\mathcal{L}_{vr}\) yields better motion predictions. Although \(\mathcal{L}_{c}\) alone do not provide quantitative gains, its combination with the right-hand–focused losses (\(\mathcal{L}_{rer}\) and \(\mathcal{L}_{vr}\)) substantially improves both body-level and right-hand accuracy. 

The next section offers a deeper examination of how these individual losses specifically enhance the quality of right-hand trajectories.

Finally, we assess \textit{IntentMotion Classifier}’s performance. As reported in Table \ref{table_classifier}, incorporating the cross-entropy loss \(\mathcal{L}_{ce}\) alongside \(\mathcal{L}_{re}\) and \(\mathcal{L}_{v}\) proposed by \cite{guo2023back} notably boosts classification f1-score. This improvement likely stems from the model’s ability to capture key motion characteristics, such as velocity and joint positioning, that are crucial for recognizing human intention. Furthermore, we evaluate two data-processing strategies for the classifier input and find that averaging the temporal dimension leads to slightly higher performance than flattening data.

\begin{table}[t]
\begin{center}
\begin{tabular}{| >{\centering\arraybackslash}m{10em} | >{\centering\arraybackslash}m{0.6cm} | >{\centering\arraybackslash}m{1.1cm} | >{\centering\arraybackslash}m{1.1cm} | >{\centering\arraybackslash}m{0.6cm} |}
\hline
\textbf{Model} & Body $L_2$ (m) & $\leq$0.20m (\%) & $\leq$0.30m (\%) & Right Hand $L_2$ (m) \\
\hline
siMLPe \cite{guo2023back} & 0.177 & 64.53 & 93.48 & 0.217 \\
\hline
IntentMotion FT & 0.171 & 69.11 & 95.27 & 0.215 \\
\hline
IntentMotion FT + Int & 0.168 & \textbf{71.66} & 97.68 & 0.200 \\
\hline
IntentMotion FT + Int + \(\mathcal{L}_{c}\) & 0.175 & 65.94 & 96.58 & 0.237 \\
\hline
IntentMotion FT + Int + \(\mathcal{L}_{c}\) + \(\mathcal{L}_{vr}\) & 0.172 & 64.32 & 95.30 & 0.205 \\
\hline
IntentMotion FT + Int + \(\mathcal{L}_{c}\) + \(\mathcal{L}_{vr}\) + \(\mathcal{L}_{rer}\) & \textbf{0.165} & 70.66 & \textbf{98.36} & \textbf{0.195} \\
\hline
\end{tabular}
\caption{IntentMotion results on leave-one-out evaluation}\vspace{-\baselineskip}
\vspace{-10pt}
\label{table_ablation}
\end{center}
\end{table}

\begin{table}[t]
\begin{center}
\begin{tabular}{| >{\centering\arraybackslash}m{17em} | >{\centering\arraybackslash}m{0.6cm} | >{\centering\arraybackslash}m{1.5cm} |}
\hline
\textbf{Model} & F1 macro & Accuracy (\%) \\
\hline
Classifier input flat. + \(\mathcal{L}_{ce}\)& 0.417 & 63.94\\
\hline
Classifier input flat. + \(\mathcal{L}_{ce}\) + \(\mathcal{L}_{re}\) + \(\mathcal{L}_{v}\) & 0.633 & 79.40 \\
\hline
Classifier input avg. + \(\mathcal{L}_{ce}\) + \(\mathcal{L}_{re}\) + \(\mathcal{L}_{v}\) & \textbf{0.654} & \textbf{81.10}  \\
\hline
\end{tabular}
\caption{Intention Classifier results on leave-one-out evaluation}
\vspace{-30pt}
\label{table_classifier}
\end{center}
\end{table}

\subsection{Qualitative Results}
In addition to the improvements quantified in our ablation studies, we emphasize that standard metrics may not fully capture the nuances of model performance. Therefore, we include visualizations to elucidate how our architectural modifications and changes in the objective function impact the generated predictions.  

As illustrated in Figure \ref{fig:figure1}, our model dynamically adjusts its predictions based on the inferred intention. When a collaborative intention is detected, the predicted motion depicts a prototypical handover scenario, characterized by the human approaching the robot and elevating the right hand to deliver the object. Conversely, when a non-collaborative intention is forced from the same input data, the prediction reflects a more reserved behavior, where the body halts its forward progression earlier and the right hand remains in a lower position. These visualizations clearly demonstrate that our architecture effectively incorporates intention as contextual information, leading to different predictions even when provided with the same input motion data.  

Moreover, the introduction of the loss terms \(\mathcal{L}_{c}\), \(\mathcal{L}_{rer}\), and \(\mathcal{L}_{vr}\) specifically enhances the modeling of right-hand dynamics. Visual comparisons indicate that \(\mathcal{L}_{c}\) helps the model better distinguish between intention-driven outputs. Additionally, the combined use of \(\mathcal{L}_{c}\), \(\mathcal{L}_{rer}\), and \(\mathcal{L}_{vr}\) results in significantly smoother and more realistic right-hand trajectories. In the absence of \(\mathcal{L}_{rer}\) and \(\mathcal{L}_{vr}\), the predicted right-hand movement appears more abrupt, underscoring the efficacy of including them in this task.

\section{Conclusions and Future Work}


In this work, we proposed a lightweight and efficient approach for human motion prediction in human-robot handovers, demonstrating that \textit{IntentMotion} achieves state-of-the-art performance while drastically improving real-time feasibility. By integrating human intention as contextual information, we enhanced both prediction accuracy and robustness, allowing for more natural and adaptive human-robot interactions. Furthermore, we introduced task-specific loss terms designed to capture handover-specific motion patterns, reinforcing the model’s ability to generalize effectively across various scenarios. 

A key contribution of this study is the development of an intention classifier within the \textit{IntentMotion} framework, enabling automatic intent recognition and linking human intention to motion prediction. This enhances forecasting accuracy and facilitates seamless collaboration by allowing robots to anticipate human actions more effectively.

Our findings emphasize the potential of lightweight, context-aware architectures for human-robot collaboration. By reducing computational complexity while maintaining accuracy, our approach supports real-time deployment in industrial automation, assistive robotics, and service robots. Future research will refine intention recognition, explore hybrid architectures integrating diverse contextual cues, and extend the method to broader HRC applications beyond handover tasks, advancing intelligent and human-centric robotic systems.

\addtolength{\textheight}{-12cm}   


\newpage
\balance
\bibliographystyle{IEEEtran}
\bibliography{bibtext/bib/references}

\end{document}